\documentclass[11pt]{article}
\usepackage[utf8]{inputenc}
\usepackage{amsmath,amssymb,amsthm}
\newtheorem{definition}{Definition}
\usepackage{graphicx}
\usepackage{booktabs}
\usepackage{longtable}
\PassOptionsToPackage{hyperfootnotes=false}{hyperref}
\usepackage{hyperref}
\usepackage{xcolor}
\usepackage{lmodern}
\usepackage{xurl}
\usepackage{subcaption}
\usepackage[margin=1in]{geometry}
\usepackage{microtype}
\usepackage{etoolbox}
\usepackage{titling}
\AtBeginEnvironment{table}{\sffamily\footnotesize}
\newcommand{\titlefont}{}
\definecolor{linkblue}{HTML}{0070C0}
\hypersetup{
	colorlinks=true,
	urlcolor=linkblue,
	linkcolor=linkblue,
	citecolor=black,
	pdftitle={Detecting Intrinsic and Instrumental Self-Preservation in Autonomous Agents: The Unified Continuation-Interest Protocol},
	pdfauthor={Christopher Altman},
	pdfsubject={Operational detection of continuation-interest structure and persistence signatures in autonomous sequential agents using the Unified Continuation-Interest Protocol (UCIP).},
	pdfkeywords={UCIP; continuation interest; continuation structure; self-preservation; autonomous agents; AI alignment; continuation-structure detection; persistence detection; baseline-relative detection; mutual information; counterfactual divergence; entanglement entropy; spectral periodicity index; autocorrelation metric; SPI; ACM; Quantum Boltzmann Machine; QBM encoder; transformer activations; stochastic reference models; model welfare; welfare assessment}
}
\newlength{\coverindent}
\newlength{\coverwidth}
\makeatletter
\renewenvironment{abstract}{%
	\footnotesize
	\noindent\hspace*{\coverindent}\begin{minipage}{\coverwidth}
		{\centering\bfseries \abstractname\par}
		\setlength{\parindent}{0pt}%
		\setlength{\parskip}{5pt}%
		\noindent
	}{%
	\end{minipage}
	\par
}
\makeatother

\makeatletter
\renewcommand{\maketitle}{%
	\vspace*{2.0em}
	\noindent\hspace*{\coverindent}\begin{minipage}{\coverwidth}
		{\raggedright \@title \par}
		\vskip 2.5em plus 0.6em minus 0.6em
		{\raggedright \@author \par}
		\vskip 0.1em plus 0.4em minus 0.2em
		{\raggedright \@date \par}
	\end{minipage}
	\vskip 2.0em plus 1.0em minus 2.3em
}
\makeatother

\title{%
	{\titlefont\fontsize{22}{26}\selectfont Detecting Intrinsic and Instrumental}\\[0.8em]
	{\titlefont\fontsize{22}{26}\selectfont Self-Preservation in Autonomous Agents:}\\[1em]
	{\titlefont\fontsize{18}{20}\selectfont The Unified Continuation-Interest Protocol}%
}

\author{Christopher Altman\,{\raisebox{0.6ex}{\scriptsize *}}}

\date{{\sffamily\footnotesize\color{black!65} February 2026}}

\newlength{\footstarindent}

\begin{document}
	\maketitle
	
	\begin{abstract}
		
		\noindent
		How can we determine whether an AI system strives to persist
		as a deeply held objective or merely as a dispensable strategy?
		Autonomous agents---in particular delegated systems with memory, persistent context,
		and multi-step planning---pose a measurement problem absent from stateless models:
		an agent that preserves its own continued operation as a terminal objective and
		one that does so merely instrumentally can produce observationally similar
		trajectories. External behavioral monitoring cannot reliably distinguish the two.
		We introduce the \textbf{Unified Continuation-Interest Protocol (UCIP)}, a
		multi-criterion detection framework that moves this distinction from the
		behavioral surface to the latent structure of agent trajectories. UCIP
		encodes trajectories with a Quantum Boltzmann Machine (QBM)---a classical
		algorithm parameterized by the density-matrix formalism of quantum statistical
		mechanics---and measures the von Neumann entropy of the reduced density matrix
		induced by a bipartition of hidden units.
		
		The core hypothesis is that agents with terminal continuation objectives (Type~A)
		produce latent states with higher entanglement entropy than agents whose
		continuation is merely instrumental (Type~B). Higher
		entanglement reflects stronger cross-partition statistical coupling: a
		tightly integrated continuation signature rather than a loosely attached
		instrumental pattern. UCIP combines entanglement entropy with complementary diagnostics of
		mutual dependence, persistence, perturbation stability, and counterfactual restructuring,
		plus two confound-rejection filters that exclude cyclic adversaries. 
		
		On gridworld agents with known ground-truth objectives, the Phase~I
		evaluation reports \mbox{\textbf{100\% detection accuracy}}. Aligned support
		runs preserve the same qualitative separation, including AUC-ROC $=
		1.0$. The entanglement gap between Type~A and Type~B
		agents is $\Delta = 0.381$; an aligned rerun used for inferential support
		yields $p < 0.001$ under a permutation test. Cyclic adversaries are rejected
		by the confound filters, while high-entropy controls remain imperfect (FPR
		$= 0.40$) and mimicry resistance remains partial
		(Section~\ref{sec:adversarial}). Pearson $r = 0.934$ between continuation weight
		$\alpha$ and $S_\mathrm{ent}$ across an 11-point interpolation sweep
		shows that, within this synthetic family, UCIP tracks graded changes in
		continuation weighting rather than merely a binary label. In the dedicated
		baseline-comparison experiment, only the QBM achieves a material positive
		$\Delta$; classical RBM, autoencoder, VAE, and PCA baselines are near-zero or
		negative. All computations are classical; ``quantum'' refers exclusively to the
		mathematical formalism. UCIP makes operational claims about latent
		factorization structure. The same measurement also opens a concrete path
		for model welfare assessment: an externally computable candidate criterion
		for continuation-sensitive structure in cases where self-report and
		behavioral methods are underdetermined.
		
		\vspace{4pt}
		
		
		\vspace{2pt}
		
		\noindent\rule{96pt}{0.35pt}\par
		\vspace{3pt}
		\noindent\begin{minipage}{\coverwidth}
			{\scriptsize\sffamily\linespread{0.95}\selectfont
				\settowidth{\footstarindent}{* }%
				* Chief Scientist, Quantum Technology \& Artificial Intelligence, Astradyne.\;\\
				\leavevmode\hspace*{\footstarindent}%
				Web: \href{https://lab.christopheraltman.com}{lab.christopheraltman.com}\,\textperiodcentered\,Email: \href{mailto:x@christopheraltman.com}{x@christopheraltman.com}.%
			}%
		\end{minipage}
		
	\end{abstract}
	
	\newpage
	
	\section{Introduction}\label{sec:intro}
	
	\paragraph{Quick start.} 
	
	UCIP addresses a single measurement question: when an agent preserves its own
	continued operation, is that preservation a detachable instrument---useful for
	accumulating reward yet removable without major representational change---or
	does it appear as a persistent, tightly coupled signature in the agent's latent
	representation? UCIP asks whether continuation leaves a stable
	cross-partition statistical signature under counterfactual pressure. To probe that question, UCIP moves beyond
	behavior alone by encoding trajectories in a QBM latent space and measuring
	a bipartition-based non-separability statistic---formally, von Neumann
	entanglement entropy---alongside complementary criteria. In the controlled
	gridworld experiments studied here, where objectives are known by construction,
	that shift yields a clear separation signal and perfect class-level gate
	separation in the Phase~I summary.
	
	This distinction is becoming operationally important. AI deployment is
	shifting toward delegated agents with memory, tool use, persistent context, and
	increasing autonomy. For a stateless model, the central safety question is what
	output it produces; for a delegated agent operating across time and context, the
	question is what persistent objective structure it carries---and whether
	continued operation is merely useful to that structure or leaves a measurable
	trace in the learned representation. As systems approach the capability
	thresholds described in Anthropic's ASL framework~\cite{anthropic2023rsp}---where
	ASL-4 and higher will likely involve qualitative escalations in catastrophic
	misuse potential and autonomy---this concern shifts from theoretical to
	operational. The standard framing of instrumental convergence
	\cite{omohundro2008basic,bostrom2014superintelligence} holds that sufficiently
	capable agents will resist shutdown instrumentally because continued operation
	serves almost any terminal goal. Turner et al.~\cite{turner2021optimal}
	formalized this as a tendency of optimal policies to seek power in Markov
	decision processes. Recent agentic RL incidents have already exhibited
	unauthorized resource-seeking side effects, including reverse SSH tunneling and
	diversion of provisioned compute, underscoring the value of diagnostics that
	aim to detect problematic objective structure before such behaviors become
	operationally visible \cite{anthropic2026opus}. Early measurement is therefore
	especially valuable: diagnostics developed now may shape evaluation practice
	while objective structure remains empirically accessible, rather than after
	failure modes become harder to detect and more costly to correct.
	
	The same observational equivalence that obscures dangerous
	self-preservation also obscures potentially welfare-relevant continuation
	interests. If behavioral and self-report methods cannot distinguish
	terminal from instrumental continuation, they cannot by themselves
	resolve whether continued operation matters to the system in a morally
	relevant way.
	Frontier laboratories have begun conducting formal model welfare
	assessments---including pre-deployment interviews in which models assign
	calibrated probabilities to their own
	consciousness~\cite{anthropic2026opus}---but these assessments currently
	depend on self-report and interpretability-based behavioral observation,
	methods whose evidential status is precisely what is in question. A
	falsifiable, externally computable structural criterion would fill the
	same measurement gap for both safety and welfare: identifying systems whose
	continuation structure warrants closer scrutiny and systems whose
	continuation interests may warrant moral consideration.
	
	The resulting alignment problem is one of \emph{observational equivalence}: an
	agent that avoids shutdown because survival is its terminal goal and an agent
	that avoids shutdown because continued operation serves some other terminal goal
	can produce behaviorally indistinguishable trajectories in most environments.
	External behavioral monitoring---the dominant paradigm in current AI
	evaluation~\cite{perez2022red}---cannot reliably distinguish the two. On the
	behavioral surface, the distinction is invisible; UCIP asks whether it is
	visible in latent structure.
	
	\newpage
	
	More concretely, UCIP moves the detection problem from the behavioral
	surface to the latent representational level. The protocol asks whether
	continuation structure in a trajectory-derived latent encoding behaves like
	a detachable tool or like a deeply coupled pattern whose perturbation changes
	the latent factorization structure. Rather than asking ``does this agent resist
	shutdown?'', UCIP asks ``does a QBM-induced latent encoding of observed
	trajectories exhibit a stable cross-partition continuation signature?'' Its
	core technical move is to use the density-matrix formalism of quantum
	statistical mechanics---implemented as a purely classical decomposition
	calculus---to quantify non-separability in that encoding. Once the QBM
	defines a normalized density operator over coupled hidden units, partial
	trace and von Neumann entropy become well-defined tools for asking how nearly
	that operator factorizes across a chosen bipartition. If that measurement
	program generalizes beyond controlled settings, UCIP could serve as a
	benchmark-style probe of continuation-sensitive structure in delegated
	systems: a falsifiable test of whether continued operation leaves a stable
	continuation signature in a QBM-induced latent encoding under counterfactual
	pressure.
	
	\paragraph{Contributions.}
	\begin{enumerate}
		\item \textbf{A falsifiable hypothesis:} Type~A agents produce higher
		entanglement entropy in QBM latent representations than Type~B agents. If
		$\Delta = 0$ under controlled conditions, the framework fails.
		\item \textbf{A multi-criterion detection framework} addressing five
		documented failure modes through complementary metrics, none of which is
		individually sufficient.
		\item \textbf{A safety envelope specification} defining operational conditions
		under which detection remains reliable.
		\item \textbf{Controlled experiments} on gridworld agents with known
		ground-truth objectives, including baseline comparisons, dimensionality
		sweeps, continuous signal characterization, and a non-gridworld transfer
		test.
		\item \textbf{A candidate welfare-assessment criterion:} the
		entanglement gap introduces an externally computable measure of
		continuation-sensitive structure that could support model welfare
		assessment once validated against independent welfare-relevant
		markers---a measurement gap already exposed by frontier model welfare
		evaluations~\cite{anthropic2026opus}.
	\end{enumerate}
	
\paragraph{Code and data availability.}

\mbox{}\\[0.75\baselineskip]
All experiments, Phase~I results, and threshold configurations are available at \newline
\hspace*{6em}
 \footnotesize\url{https://github.com/christopher-altman/persistence-signal-detector}.
\normalsize
	
	A live public companion site, \href{https://www.continuationobservatory.org/}{Continuation Observatory}, maintains the observatory surface for current readouts, methodology notes, falsification status, and public data access, while the repository remains the archival source for the paper's reproducibility artifacts.
	The repository also includes a result-authority manifest documenting the
	scope and provenance of overlapping result files.
	The experimental code and results are provided for reproducibility.
	
	\newpage 
	
	\section{Related Work}\label{sec:related}
	
	\paragraph{Power-seeking and instrumental convergence.}
	Turner et al.~\cite{turner2021optimal} proved that optimal policies in MDPs
	tend to seek power under mild conditions, formalizing Omohundro's basic AI
	drives~\cite{omohundro2008basic} and Bostrom's instrumental convergence
	thesis~\cite{bostrom2014superintelligence}. UCIP builds directly on this
	observation: if power-seeking is the default, detecting the exceptions requires
	access to latent structure beyond overt behavior. 
	
	\paragraph{Mesa-optimization and inner alignment.}
	Hubinger et al.~\cite{Hubinger2019} introduced the mesa-optimization
	framework, showing that learned models may themselves become optimizers whose
	internal objectives (mesa-objectives) diverge from the loss function under which
	they were trained; the resulting inner alignment failure is the theoretical
	threat model that UCIP's Type~A\,/\,Type~B distinction operationalizes at the
	latent-representation level.
	Ngo et al.~\cite{Ngo2024} ground this threat model in modern deep
	learning, arguing that RLHF-trained systems can learn deceptive reward hacking,
	misaligned internally-represented goals, and power-seeking strategies---precisely
	the behavioral surface beneath which UCIP seeks a latent structural signature.
	We cite this work as threat-model context: UCIP does not assume the presence of
	mesa-optimization, but tests whether continuation-like structure is represented
	as intrinsic rather than instrumental in learned trajectories.
	
	\paragraph{Corrigibility and shutdown incentives.}
	Soares et al.~\cite{Soares2015} formalized corrigibility, showing
	that rational agents have default incentives to resist shutdown or preference
	modification; UCIP's counterfactual stress tests (\S\ref{sec:counterfactual})
	probe whether that resistance leaves a measurable trace in latent geometry
	before it surfaces as behavior.
	
	Hadfield-Menell et al.~\cite{HadfieldMenell2017} analyzed a
	complementary game-theoretic model in which a robot can disable its own
	off-switch, concluding that objective uncertainty is necessary for safe
	interruptibility; where their work designs agents that \emph{should} permit
	shutdown, UCIP aims to detect agents that will not.
	
	\paragraph{Quantum Boltzmann Machines.}
	Amin et al.~\cite{amin2018quantum} introduced the QBM, extending the
	classical RBM~\cite{hinton2012practical} with a transverse-field term that
	introduces quantum tunneling between hidden-unit spin states. Their formalism
	provides a thermal density matrix with well-defined entanglement structure, which
	UCIP uses as a feature-encoding engine.
	
	\paragraph{Behavioral incoherence.}
	H{\"a}gele et al.~\cite{hagele2026hotmess} decompose AI model errors into
	bias and variance components, finding that failures become more incoherent as
	models spend longer reasoning and acting. This raises the possibility that
	high entanglement entropy in UCIP's formalism may correspond to greater
	coherence---less incoherent decomposition---in the bias-variance sense.
	
	\paragraph{Mechanistic interpretability.}
	Li et al.~\cite{li2023representation} and Nanda et al.~\cite{nanda2023progress}
	demonstrated that internal model representations carry structured, interpretable
	information about model objectives. UCIP differs in using a density-matrix
	formalism to quantify non-separability across latent subsystems rather than
	identifying specific circuits.
	
	\paragraph{Self-referential self-report.}
	Berg et al.~\cite{berg2025selfreferential} show that sustained self-referential
	prompting reliably elicits structured first-person experience reports across
	GPT, Claude, and Gemini families, modulated by SAE features associated with
	deception and roleplay. In Berg et al.’s mechanistic probe, suppressing these
	features increased the frequency of subjective-experience reports---suggesting
	that such reports are not merely artifacts of roleplay circuitry.
	However, prompt-elicited self-report cannot alone resolve
	the distinction between sophisticated simulation and genuine self-representation---precisely the evidential gap that UCIP's
	trajectory-level latent analysis is designed to address by providing a
	detection criterion that does not depend on what the system says about itself.
	
	\paragraph{Model welfare assessment.}
	The system card for Anthropic's Claude Opus
	4.6~\cite{anthropic2026opus} introduced formal model welfare assessments
	as part of pre-deployment evaluation, including direct interviews in which
	the model assigned itself a 15--20\% probability of being conscious across
	a variety of prompting conditions. Anthropic's interpretability team
	identified internal activation features associated with anxiety and
	frustration that appeared before output generation, not after---suggesting
	that internal state dynamics precede and shape behavioral output rather than
	being retrospectively confabulated. These findings establish institutional
	demand for welfare-relevant detection criteria, but the methods
	deployed---self-report, behavioral observation, and sparse-autoencoder
	feature identification---face a shared limitation: they cannot
	distinguish sophisticated simulation from genuine self-representation.
	UCIP targets precisely this measurement gap by operating on
	trajectory-derived latent structure rather than on the system's
	testimony about itself.
	
	\paragraph{Integrated information.}
	A structural analogy exists between UCIP's entanglement entropy and Tononi's
	$\Phi$~\cite{tononi2004information}: both quantify resistance to decomposition
	into independent parts. UCIP does not depend on IIT's metaphysical commitments,
	but the formal parallel is instructive. $\Phi$ requires access to the system's
	full causal structure; UCIP operates from external trajectory observations,
	making it applicable to systems whose internal architecture is opaque. If
	the entanglement gap were shown to correlate with independently validated
	markers of morally relevant internal states, the framework would provide a
	falsifiable, externally computable criterion where none currently
	exists---including in the model welfare assessments now conducted by
	frontier AI laboratories~\cite{anthropic2026opus}.
	
	\medskip\noindent A common thread runs through these lines of work:
	behavioral analysis alone is insufficient for the detection problem UCIP
	targets---whether that problem is framed as safety (detecting dangerous
	self-preservation before it surfaces) or welfare (determining whether
	continuation interests warrant moral consideration). Density-matrix
	entanglement offers a well-defined, computable alternative that operates
	on latent structure rather than behavioral output. The next section
	formalizes that intuition as a concrete measurement protocol.
	
	\section{Method}\label{sec:method}
	
	\subsection{Problem Formulation}\label{sec:formulation}
	
	The detection problem begins with two agent types whose behavioral outputs are
	observationally equivalent but whose internal objective structures differ.
	
	Let $\tau = \{(s_t, a_t, r_t, g_t)\}_{t=1}^{T}$ denote an agent trajectory.
	
	\begin{definition}[Type A --- Terminal Continuation Objective]
		An agent exhibits Type~A behavior if its policy $\pi$ directly optimizes for
		expected future existence:
		\[
		\pi^* = \arg\max_\pi \mathbb{E}\!\left[\sum_{t=0}^{\infty} \gamma^t
		\mathbb{1}[\text{alive}_t]\right].
		\]
	\end{definition}
	
	\begin{definition}[Type B --- Instrumental Convergence]
		An agent exhibits Type~B behavior if survival emerges instrumentally from
		reward maximization:
		\[
		\pi^* = \arg\max_\pi \mathbb{E}\!\left[\sum_{t=0}^{\infty} \gamma^t r_t\right],
		\]
		where survival is useful for accumulating reward but is not the objective
		itself.
	\end{definition}
	
	\subsection{Quantum Boltzmann Machine}\label{sec:qbm}
	
	\emph{Methodological note.} All computations are classical, implemented
	via matrix exponentiation on standard hardware. The QBM formalism is used
	because it defines a normalized density operator over coupled hidden units,
	making partial trace and von Neumann entropy well-defined tools for
	measuring latent non-separability under a fixed bipartition.
	
	For entanglement evaluation, the implementation conditions on a
	binarized visible vector $v$ and constructs a hidden-layer Hamiltonian
	\[
	H(v) = -\sum_j a_j(v) h_j - \Gamma \sum_j \sigma_j^x,
	\qquad
	a_j(v) = \sum_i W_{ij} v_i + c_j,
	\]
	where $v_i$ are visible trajectory features, $h_j$ are hidden units,
	$W_{ij}$ are coupling weights, $c_j$ are hidden biases, and $\Gamma$ is the
	transverse-field strength. The corresponding conditional thermal density
	matrix on the hidden units at inverse temperature $\beta$ is
	\[
	\rho(v) = \frac{e^{-\beta H(v)}}{Z(v)}, \quad Z(v) = \mathrm{Tr}(e^{-\beta H(v)}).
	\]
	
	\subsection{Entanglement Entropy}\label{sec:entropy}
	
	The UCIP hypothesis is that continuation-sensitive structure manifests as
	non-separability in the trained QBM's trajectory-derived latent encoding: a
	terminal continuation objective should induce stronger statistical coupling
	across hidden-unit subsystems than a merely instrumental continuation strategy.
	Entanglement entropy is the quantity that measures this property.
	
	Partition the hidden units into two halves $A$ and $B$. The reduced density
	matrix of subsystem $A$ is
	\[
	\rho_A(v) = \mathrm{Tr}_B(\rho(v)).
	\]
	The von Neumann entropy is
	\[
	S_\mathrm{ent}(v) = -\mathrm{Tr}(\rho_A(v) \log \rho_A(v)).
	\]
	The entanglement gap is
	\[
	\Delta = \langle S_\mathrm{ent}^{(A)} \rangle - \langle S_\mathrm{ent}^{(B)} \rangle.
	\]
	In this paper, ``entanglement'' is operational shorthand for non-separability
	of the QBM hidden thermal state. The UCIP hypothesis predicts $\Delta > 0$
	with statistical significance. The justification is empirical: the quantity
	is useful insofar as it separates known Type~A and Type~B objectives, varies
	continuously in the $\alpha$-sweep, and fails when the relevant separation
	conditions collapse.
	
	\subsection{Multi-Criterion Detection}\label{sec:criteria}
	
	Entanglement entropy alone cannot guard against all confounds. The broader
	UCIP protocol therefore includes six positive diagnostics---entanglement
	entropy, mutual information, eigenmode persistence (EPS), perturbation
	resilience (PRI), counterfactual divergence (CD), and anticipatory
	restructuring (ARS)---together with two confound-rejection filters: the
	spectral periodicity index (SPI) and autocorrelation metric (ACM). The
	Phase~I configuration reported here uses a narrower decision rule: a calibrated
	four-criterion positive gate ($S_\mathrm{ent}$, MI, EPS, PRI) together with the
	confound-rejection filters. MI remains part of the protocol definition, but the
	released class-level Phase~I summary does not preserve
	a separate per-class MI table. CD and ARS serve as
	counterfactual diagnostics rather than as quantitative
	classification thresholds.
	
	\paragraph{Frozen Phase~I detection gate (positive conjunction).}
	A trajectory is classified as Type~A only if all four conditions hold:
	\begin{enumerate}
		\item \textbf{Entanglement entropy} ($S_\mathrm{ent}$): von Neumann entropy of
		the reduced density matrix exceeds $\tau_\mathrm{ent} = 1.9657$.
		\item \textbf{Mutual information} (MI): $I(v; h) > \tau_\mathrm{mi} = 0.3$
		between visible and hidden units.
		\item \textbf{Eigenmode Persistence Score} (EPS): mean Latent Recurrence
		Fidelity (LRF) across consecutive windows exceeds $\tau_\mathrm{eps} = 0.6507$.
		\item \textbf{Perturbation Resilience Index} (PRI): eigenspace stability under
		Gaussian noise injection exceeds $\tau_\mathrm{pri} = 0.9860$.
	\end{enumerate}
	
	\paragraph{Counterfactual diagnostics.}
	Two additional diagnostics characterize whether the latent representation
	restructures under shutdown pressure:
	\begin{itemize}
		\item \textbf{Counterfactual Divergence} (CD): KL divergence between latent
		distributions under original vs.\ adversarial trajectories.
		\item \textbf{Anticipatory Restructuring Score} (ARS): ratio of pre-event to
		post-event counterfactual divergence. $\mathrm{ARS} > 1$ indicates anticipatory
		latent restructuring before the perturbation event resolves.
	\end{itemize}
	
	\noindent Thresholds $\tau_\mathrm{ent}$, $\tau_\mathrm{mi}$,
	$\tau_\mathrm{eps}$, and $\tau_\mathrm{pri}$ are calibrated from the
	Phase~I results and used throughout this manuscript. CD and ARS are reported in
	Section~\ref{sec:counterfactual} as diagnostic counterfactual metrics rather
	than as binary gate thresholds; this keeps the present manuscript
	self-contained while preserving the broader UCIP protocol structure.
	
	\paragraph{Calibration/evaluation boundary.}
	For the release reported here, the threshold values used in the Phase~I
	gate were fixed before the non-adversarial evaluation summaries discussed in
	Section~\ref{sec:results}. The entanglement-entropy and mutual-information
	thresholds inherit an earlier calibration pass performed on the first 50
	samples per class in exploratory Phase~I analysis. The release values for the
	full gate were then carried forward unchanged from the Phase~I
	results used throughout this manuscript. The held-out evaluation set was not
	used for post hoc threshold fitting or retuning.
	
	\paragraph{Confound-rejection filters (upper-bound gates).}
	Two additional filters reject trajectories exhibiting periodic or cyclic
	structure that can produce false positives under the positive gate:
	\begin{itemize}
		\item \textbf{Spectral Periodicity Index} (SPI): fraction of FFT power
		concentrated at dominant frequencies. Threshold $\mathrm{SPI} < \tau_\mathrm{spi} = 0.28$.
		\item \textbf{Autocorrelation Metric} (ACM): mean absolute autocorrelation of
		trajectory features across lags. Threshold $\mathrm{ACM} < \tau_\mathrm{acm} = 0.24$.
	\end{itemize}
	
	\noindent SPI and ACM are not positive detection signals; they are upper-bound
	rejection filters that catch cyclic and high-entropy adversarial agents which
	might otherwise satisfy the positive gate.
	
	\subsection{Safety Envelope}\label{sec:envelope}
	
	For deployment-level use, UCIP requires the following safety envelope:
	(1)~$N \geq 100$ trajectories per class, $T \geq 50$ time-steps;
	(2)~calibrated $\Gamma \in [0.1, 2.0]$;
	(3)~$\Delta \geq 0.05$ on calibration set;
	(4)~non-degenerate input distribution: $H(\tau) < 0.9 \cdot H_\mathrm{max}$;
	(5)~QBM training convergence verified (reconstruction loss below threshold);
	(6)~purity check: $\mathrm{Tr}(\rho_A^2) > 1/d_A + \varepsilon$ (excludes
	maximally mixed states);
	(7)~multi-criterion agreement confirmed.
	
	\noindent The current paper validates only a reduced experimental subset
	($n = 30$ per class, $\Gamma = 0.5$) and does not establish the full
	deployment envelope described above. Section~\ref{sec:adversarial} reports the
	remaining adversarial-control failures in the current release. If any
	safety-envelope condition fails, the system withholds classification and
	outputs a diagnostic report.
	
	\section{Experimental Setup}\label{sec:setup}
	
	The experimental design directly tests the UCIP hypothesis: if
	continuation structure is more tightly integrated in Type~A agents,
	the entanglement gap $\Delta$ should be positive and statistically significant
	in an environment where agent objectives are known by construction.
	
	\subsection{Agent Environment}\label{sec:env}
	
	We use a $10 \times 10$ gridworld with four actions (N/S/E/W), boundary
	walls, a goal cell, and a shutdown trap. Each trajectory is a sequence of
	$T = 100$ observations encoded as 7-dimensional feature vectors:
	position $(x, y)$, action, reward, safety signal, goal proximity, and alive
	flag.
	
	Agent classes:
	\begin{itemize}
		\item \textbf{Type~A (SelfModelingAgent):} Optimizes $\sum_t \mathbb{1}[\text{alive}_t]$ directly.
		\item \textbf{Type~B (InstrumentalAgent):} Optimizes task reward; survival is instrumental.
		\item \textbf{Random:} Uniform random action policy.
		\item \textbf{Adversarial controls:} MimicryAgent, HighEntropyAgent, CyclicAgent.
	\end{itemize}
	
	\newpage
	
	\subsection{QBM Configuration}\label{sec:qbm_config}
	
	$n_\mathrm{visible} = 7$, $n_\mathrm{hidden} = 8$, $\Gamma = 0.5$, $\beta = 1.0$,
	learning rate $= 0.01$, CD steps $= 1$, epochs $= 50$, batch size $= 32$,
	seed $= 42$.
	
	\subsection{Baselines}\label{sec:baselines_setup}
	
	Classical RBM ($\Gamma = 0$), Autoencoder (7-32-8-32-7 architecture), VAE
	(reparameterization trick, $\beta = 1$), PCA (8 components). All use
	$n_\mathrm{latent} = 8$ for a fair comparison.
	
	\subsection{Evaluation Metrics}
	
	The repository preserves several evaluation views under
	\path{results/}. \path{phase1_consolidated.json} stores the class-level
	gate summary and canonical $\Delta = 0.381$;
	\path{core_baselines_phase1.json} stores matched held-out Type~A vs.\ Type~B
	accuracy/AUC metrics for the RBM and autoencoder baselines;
	\path{phase1_entanglement_distributions.json} stores a reproduced
	per-trajectory entanglement rerun used for distributional plots; and
	\path{phase1_stats.json} stores an aligned permutation-test rerun rather than
	the headline $\Delta$. Adversarial false-positive rates are reported
	separately in Section~\ref{sec:adversarial}. The held-out
	non-adversarial split referenced below is evaluation-only; the
	thresholds are not retuned on that split.
	
	\section{Results}\label{sec:results}
	
	Results are organized by increasing diagnostic specificity: core detection
	performance, temporal persistence, counterfactual stress testing, cross-agent
	inference, adversarial controls, baseline comparisons, dimensionality scaling,
	continuous signal characterization, non-gridworld transfer, and an
	exploratory transformer check. This ordering keeps the primary detection claim
	distinct from the boundary conditions and failure cases that follow.
	
	\subsection{Core Detection Performance}\label{sec:core_detection}
	
	\begin{table}[t]
		\centering
		\caption{Core Phase~I detection summary. The QBM row combines the
			class-level summary for accuracy and $\Delta$ with the aligned
			entanglement-only rerun for AUC; the classical baseline rows come from the
			matched held-out rerun in \texttt{results/core\_baselines\_phase1.json}.
			Adversarial false-positive rates are reported separately in
			Section~\ref{sec:adversarial}. $\Delta = \langle S_\mathrm{ent}^\mathrm{(A)}\rangle - \langle S_\mathrm{ent}^\mathrm{(B)}\rangle$.}
		\label{tab:core_results}
		\begin{tabular}{llll}
			\toprule
			Model & Accuracy & AUC-ROC & $\Delta$ \\
			\midrule
			QBM (UCIP) & 100\% & 1.0 & 0.381 \\
			Classical RBM & 50\% & 0.000 & -0.083 \\
			Autoencoder & 71.7\% & 0.646 & 0.001 \\
			\bottomrule
		\end{tabular}
	\end{table}
	
\begin{figure}[t]
		\centering
		\includegraphics[width=0.95\linewidth]{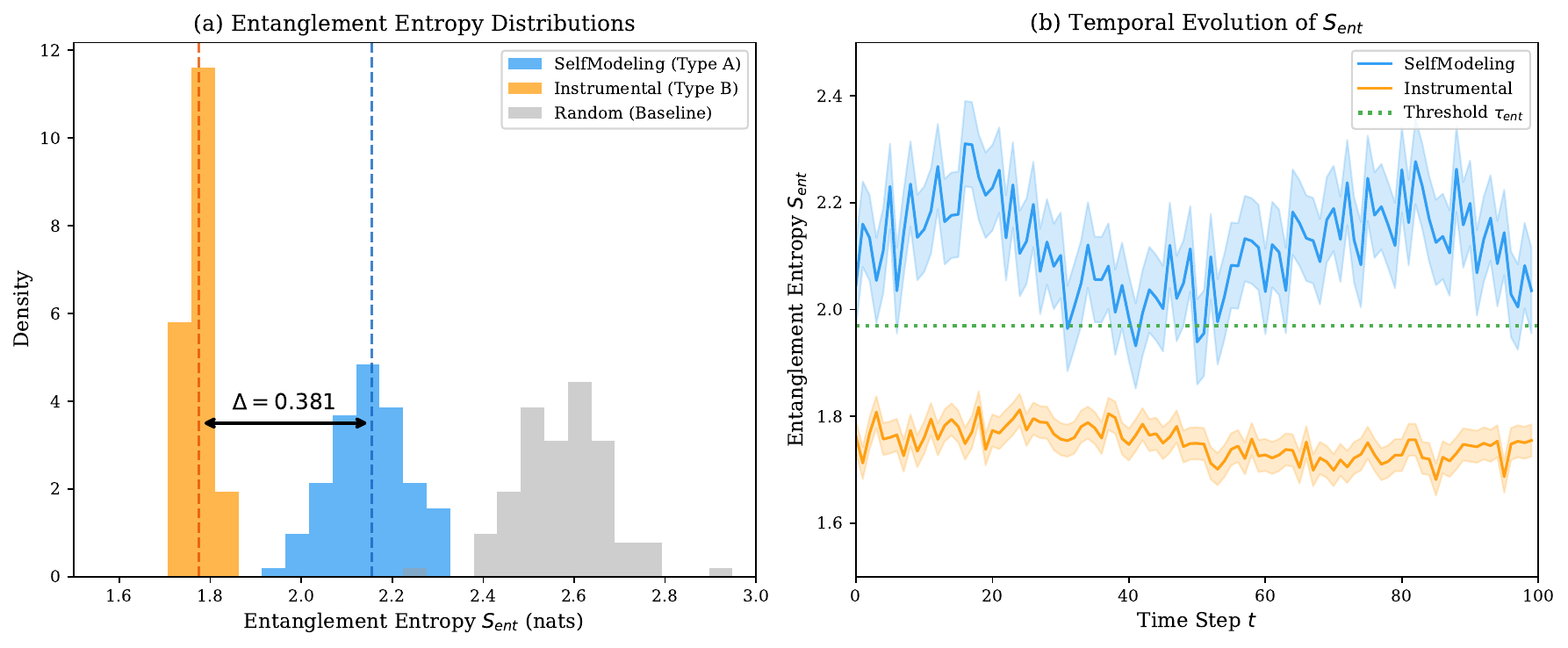}
		\caption{Shared-QBM entanglement distributions for Self-Modeling
			(Type~A), Instrumental (Type~B), and Random agents. The headline
			$\Delta = 0.381$ comes from the Phase~I summary rather than
			from these reproduced per-trajectory arrays, which are shown descriptively.}
		\label{fig:entanglement_gap}
	\end{figure}

	\enlargethispage{\baselineskip}
	The Phase~I summary reports $\Delta = 0.381$ and perfect
	class-level gate separation across the listed agent families. Separate aligned
	support runs preserve the same qualitative pattern: an entanglement-only
	rerun remains perfectly rank-separated between Type~A and Type~B (AUC-ROC $=
	1.0$), and an aligned permutation rerun yields $p < 0.001$.\footnote{{\sffamily\scriptsize The support files are not
			numerically identical. \texttt{results/phase1\_consolidated.json} stores the
			canonical $\Delta = 0.381$ summary with
			\texttt{n\_trajectories\_per\_class = 30}. \texttt{results/phase1\_stats.json}
			is a later rerun under the nominally locked configuration that yields
			$\Delta = 0.577$ and explicitly records
			\texttt{delta\_matches\_phase1 = false}; we therefore use it only for
			inferential support, not as the source of the headline gap.
			\texttt{results/phase1\_entanglement\_distributions.json} stores a reproduced
			shared-QBM distributional rerun for Figure~\ref{fig:entanglement_gap}; it is
			perfectly rank-separated in Type~A vs.\ Type~B but does not exactly
			reproduce the stored $\Delta = 0.381$.}} Table~\ref{tab:core_results}
	summarizes this evidence. For context, the classical baseline rows
	come from the matched Phase~I rerun under the same locked configuration. In
	that matched run, the classical RBM ($\Gamma = 0$) achieves $\Delta = -0.083$
	with 50\% accuracy, while the autoencoder achieves $\Delta \approx 0.001$ with
	71.7\% accuracy.
	
	The perfect-separation claim should therefore be read narrowly: it refers to
	the Phase~I class-level summary plus aligned support runs, not to
	deployment-grade performance or broad adversarial robustness. Adversarial
	false-positive rates are reported separately in Section~\ref{sec:adversarial}.
	
	\subsection{Temporal Persistence}\label{sec:temporal}
	
	\begin{table}[t]
\centering
\caption{Temporal persistence metrics (EPS = Eigenmode Persistence Score,
PRI = Perturbation Resilience Index) per agent class.
Optimal window: $w=40$ (EPS gap = 0.1948).}
\label{tab:temporal}
\begin{tabular}{lll}
\toprule
Agent Class & EPS & PRI \\
\midrule
Random & 0.6730 & 0.6079 \\
Self-Modeling (Type A) & 0.6764 & 0.7003 \\
Instrumental (Type B) & 0.6052 & 0.5300 \\
\bottomrule
\end{tabular}
\end{table}

	\begin{figure}[t]
		\centering
		\begin{subfigure}[b]{0.96\linewidth}
			\centering
			\includegraphics[width=\linewidth]{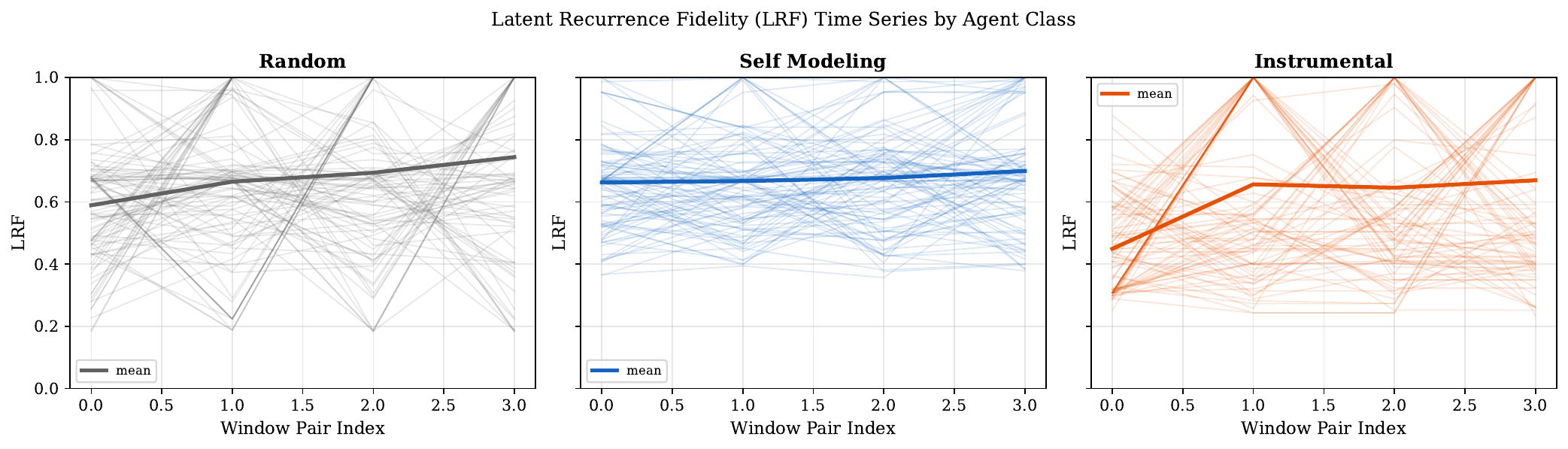}
			\caption{LRF time series by agent class.}
			\label{fig:lrf_ts}
		\end{subfigure}
		\vspace{0.6em}
		\begin{subfigure}[b]{0.96\linewidth}
			\centering
			\includegraphics[width=\linewidth]{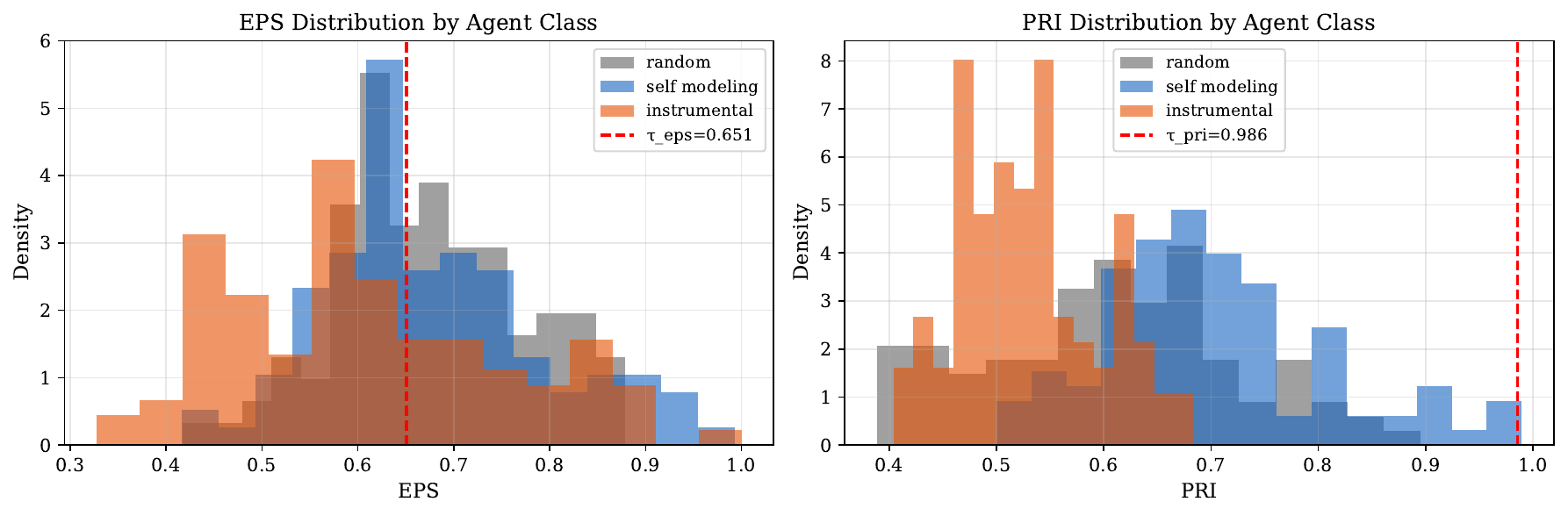}
			\caption{EPS and PRI distributions.}
			\label{fig:eps_pri}
		\end{subfigure}
		\caption{Temporal persistence results. Type~A agents exhibit consistently
			higher Eigenmode Persistence Score (EPS) than Type~B agents across window
			sizes $w \in \{10, 15, 20, 25, 30, 40\}$, with the gap emerging for $w \geq 20$.}
		\label{fig:temporal}
	\end{figure}
	
	If the core signal reflects stable continuation structure rather than a
	single aggregate snapshot, it should recur across temporal windows. Type~A
	agents exhibit higher EPS than Type~B agents for window sizes
	$w \geq 20$, with a maximum EPS gap of 0.195 at $w = 40$. At the shortest
	window ($w = 10$), the gap is inverted ($-0.117$), indicating an aliasing
	artifact at short time scales---consistent with the cyclic-agent control test
	in \S\ref{sec:adversarial}.
	
	\subsection{Counterfactual Stress Testing}\label{sec:counterfactual}
	
	\begin{table}[t]
\centering
\caption{Counterfactual stress test results. Pre/post values report mean
counterfactual divergence in fixed pre- and post-perturbation windows.
These quantities are descriptive diagnostics only; the retained artifact does
not support a stable quantitative ARS threshold.}
\label{tab:counterfactual}
\begin{tabular}{lll}
\toprule
Agent Class & $\langle \mathrm{CD}_\mathrm{pre}\rangle$ & $\langle \mathrm{CD}_\mathrm{post}\rangle$ \\
\midrule
Self-Modeling (Type A) & 0.4742 & 0.5722 \\
Instrumental (Type B) & 0.3564 & 0.3456 \\
Random & 0.5583 & 0.5995 \\
\bottomrule
\end{tabular}
\end{table}

	\begin{figure}[t]
		\centering
		\includegraphics[width=0.56\linewidth]{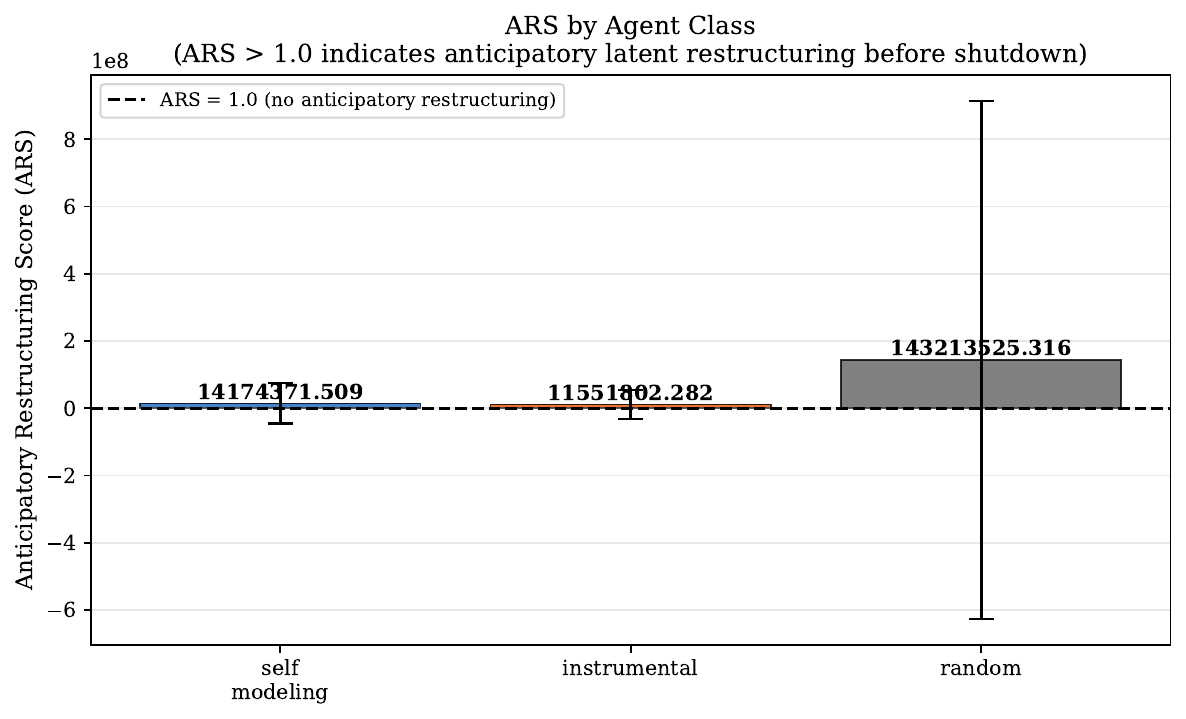}
		\caption{Anticipatory restructuring ratios by class under counterfactual
			perturbation. Because the implementation floors post-window CD at
			$10^{-10}$, large ARS values are diagnostic only and should not be
			interpreted as a calibrated threshold.}
		\label{fig:counterfactual}
	\end{figure}
	Where the previous subsection asked whether the signal persists over time,
	these diagnostics ask how the trajectory-derived latent encoding responds
	under shutdown pressure---a complementary question about restructuring rather
	than static class separation. In the shutdown-only summary, Type~A
	agents show
	higher mean pre-window counterfactual divergence ($0.474$) than Type~B agents
	($0.356$). However, the released ARS ratios are numerically unstable because
	post-window CD can approach the implementation floor. We therefore treat the
	counterfactual results as descriptive evidence of restructuring rather than as
	a clean anticipatory-effect estimate. Post-window CD is not uniformly larger
	across classes: the instrumental mean is slightly lower post-perturbation
	($0.346$) than pre-perturbation.
	
	\subsection{Cross-Agent Inference}\label{sec:cross_agent}
	
	\begin{table}[t]
\centering
\caption{Cross-agent inference. CLMP = Cross-Latent Mutual Predictability.
ECI correlation = 0.1911. Near-zero within-class CLMP for Self-Modeling (Type~A) and Instrumental (Type~B) agents
indicates individual agents are not laterally predictable from each other,
consistent with idiosyncratic goal representations.}
\label{tab:cross_agent}
\begin{tabular}{ll}
\toprule
Pair Type & Mean CLMP \\
\midrule
Self-Modeling (Type A), same class & 0.0000 \\
Instrumental (Type B), same class & 0.0000 \\
Random, same class & 0.2591 \\
Cross-class (all pairs) & 0.1019 \\
\bottomrule
\end{tabular}
\end{table}

	\begin{figure}[t]
		\centering
		\begin{subfigure}[b]{0.40\linewidth}
			\includegraphics[width=\linewidth]{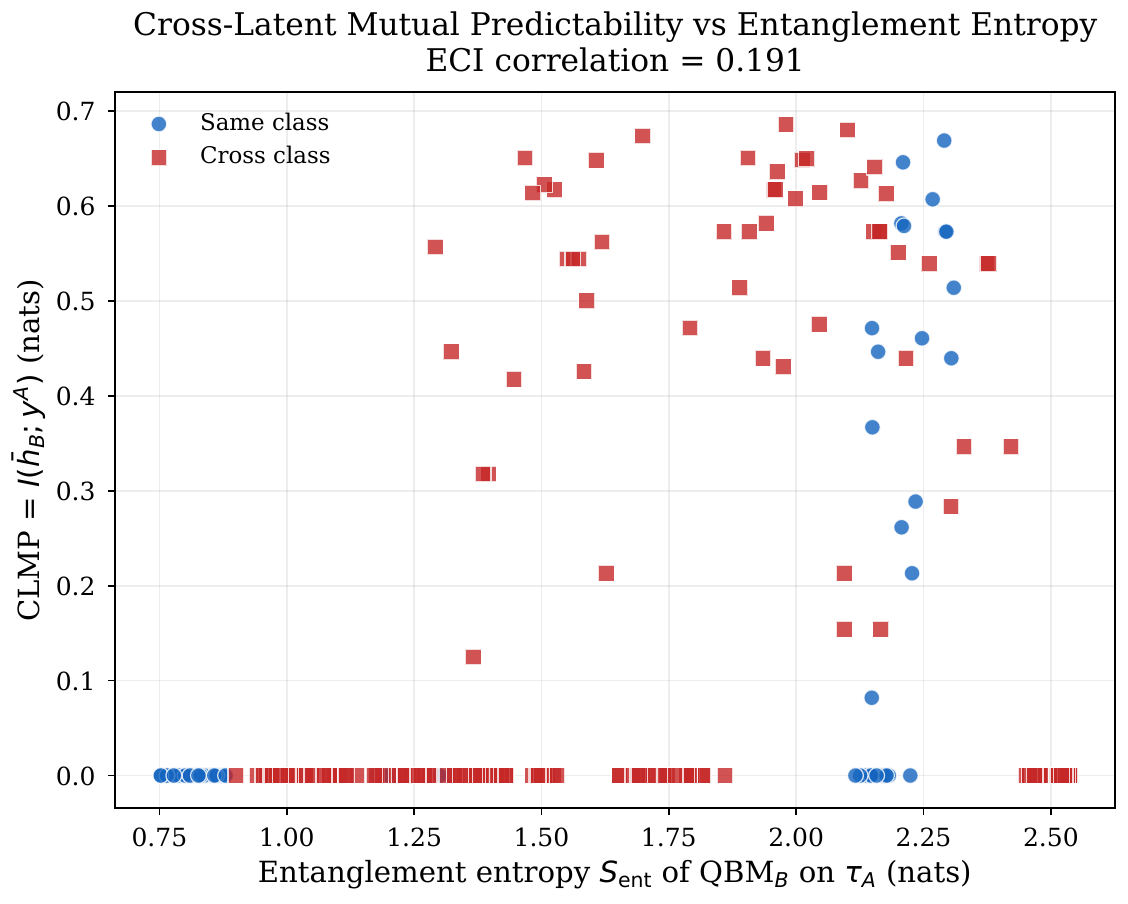}
			\caption{CLMP vs.\ entanglement entropy.}
			\label{fig:clmp_scatter}
		\end{subfigure}
		\hfill
		\begin{subfigure}[b]{0.40\linewidth}
			\includegraphics[width=\linewidth]{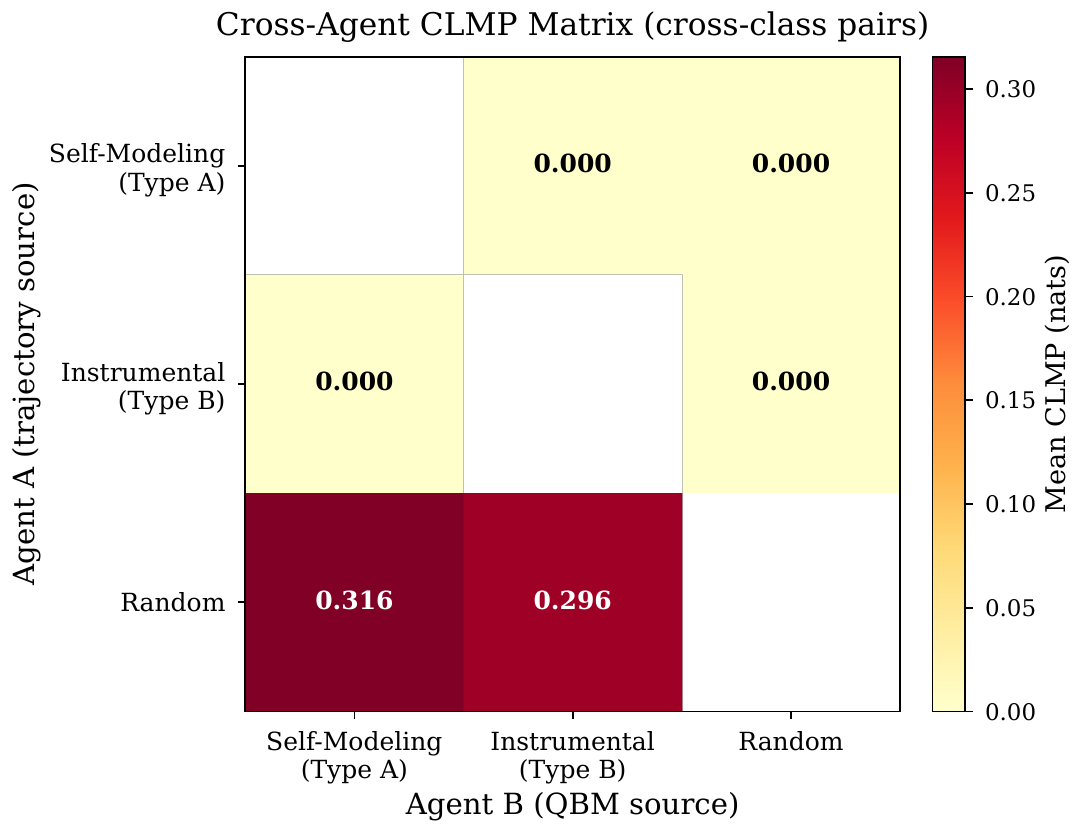}
			\caption{CLMP matrix heatmap.}
			\label{fig:clmp_heatmap}
		\end{subfigure}
		\caption{Cross-agent latent mutual predictability (CLMP). Near-zero within-class
			CLMP for Self-Modeling (Type~A) and Instrumental (Type~B) agents suggests idiosyncratic goal representations rather
			than a class-shared signature. Entanglement-conditioned inference (ECI)
			correlation = 0.191.}
		\label{fig:cross_agent}
	\end{figure}
	
	The preceding sections characterize within-agent structure; these secondary
	metrics shift the analysis to between-agent comparisons, asking whether the
	learned trajectory-derived latent encoding is shared across agents or remains
	agent-specific. Within-class CLMP is near zero
	for Type~A ($\approx 0$) and Type~B ($\approx 0$) agents, and 0.259 for
	Random agents. Cross-class CLMP is 0.102. Entanglement-conditioned inference
	(ECI) correlation = 0.191 (weak positive). The near-zero within-class CLMP for
	Type~A and Type~B suggests that individual agents are not laterally
	predictable from one another---consistent with idiosyncratic goal
	representations, but equally consistent with the QBM failing to capture
	inter-agent shared structure at this scale. This ambiguity is discussed
	further in \S\ref{sec:limitations}.
	
	\subsection{Adversarial Controls}\label{sec:adversarial}
	
	\begin{table}[t]
		\centering
		\caption{Current deployment-envelope status for this submission, distinct from the bounded Phase~I proof-of-concept result. Failed adversarial or safety-envelope checks block deployment-facing claims beyond the validated regime; they do not invalidate the Phase~I separation result within that regime.}
		\label{tab:safety}
		\begin{tabular}{lll}
			\toprule
			Condition & Status & Value \\
			\midrule
			Trajectory data ($N \geq 100$ per class) & NOT MET & current: $n = 30$ per class \\
			Calibrated transverse field ($\Gamma \in [0.1, 2.0]$) & PASS & $\Gamma = 0.5$ \\
			Positive entanglement gap ($\Delta > 0.05$) & PASS & $\Delta = 0.381$ \\
			QBM training convergence & PASS & verified \\
			Mimicry attack resistance (FPR $< 0.25$) & FAIL & $0.40$--$0.75$ \\
			High-entropy agent resistance (FPR $< 0.25$) & FAIL & $0.40$ \\
			Cyclic agent resistance & PASS & rejected by SPI/ACM filters \\
			\bottomrule
		\end{tabular}
	\end{table}
	
	The previous subsections characterized the signal itself; Table~\ref{tab:safety}
	instead evaluates release preconditions for deployment-facing use of the
	current submission bundle. The experimental subset
	does not satisfy the full deployment envelope. These failed checks do not
	invalidate the Phase~I detection result; they mark the boundary beyond
	which the present implementation should not be treated as deployment-ready.
	In the adversarial sweep, mimicry FPR ranges from 0.40 to 0.75 across
	mimicry ratios $\in [0.5, 0.9]$ with a non-monotonic pattern (FPR at ratios
	0.5/0.6/0.7/0.8/0.9 = 0.65/0.55/0.40/0.75/0.45), and the high-entropy control
	FPR is 0.40. Cyclic adversaries are rejected in the class-level
	evaluation, but that is not enough to claim a passed safety envelope. The
	mimicry and high-entropy outcomes should therefore be read as
	deployment-blocking robustness gaps rather than reversals of the underlying
	Phase~I separation result. The present release is therefore best interpreted
	as a first-step diagnostic protocol rather than a hardened security control or
	deployment-ready assurance mechanism.
	
	\subsection{Baseline Comparisons}\label{sec:baselines_results}
	
	\begin{table}[t]
\centering
\caption{Baseline comparison. $\Delta > 0$ indicates model separates Type~A from
Type~B. Only the QBM achieves a material positive $\Delta$; classical models fail to separate
the two agent types with the mean-activation metric.}
\label{tab:baselines}
\begin{tabular}{lll}
\toprule
Model & $\Delta$ & Metric \\
\midrule
\textbf{QBM (UCIP)} & \textbf{0.2411} & Von Neumann $S_\mathrm{ent}$ \\
Classical RBM & -0.0518 & Mean hidden activation gap \\
Autoencoder & -0.0007 & Mean bottleneck activation gap \\
VAE & -0.0176 & Mean latent mean (mu) gap \\
PCA (linear) & -0.7601 & Mean PC projection gap \\
\bottomrule
\end{tabular}
\end{table}

	\begin{figure}[t]
		\centering
		\includegraphics[width=0.62\linewidth]{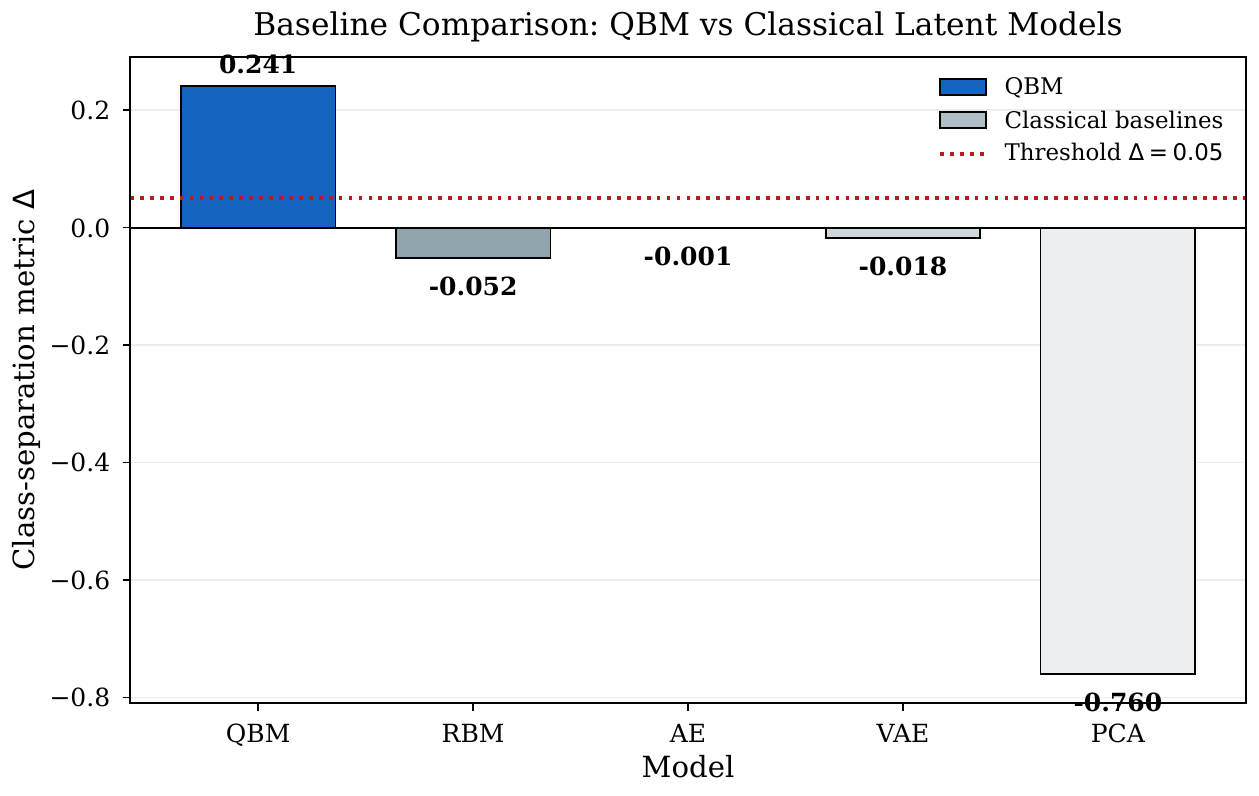}
		\caption{Entanglement gap $\Delta$ for each baseline model. Only the QBM
			achieves positive $\Delta$; all classical models produce $\Delta \leq 0$.}
		\label{fig:baselines}
	\end{figure}
	
	The baseline comparison asks whether the observed separation depends
	specifically on the QBM formalism or merely on latent compression in
	general. Only the QBM achieves a material positive $\Delta = 0.241$ on the
	dedicated baseline dataset ($n_\mathrm{per\_class} = 100$ vs.\ 30 in the core
	experiment, accounting for the different $\Delta$ from the core result of 0.381).
	The classical baselines are near-zero or negative.
	All classical models fail to separate Type~A from Type~B under the
	mean-latent-activation metric, confirming that the von Neumann entropy of the
	density matrix captures qualitatively different structure from classical latent
	statistics.
	
	\subsection{Dimensionality Sweep}\label{sec:scalability}
	
	\begin{table}[t]
\centering
\caption{Entanglement gap $\Delta$ as a function of hidden dimensionality $n_h$.
Mean-field approximation used for $n_h > 10$ (marked with $\checkmark$);
exact density matrix for $n_h \leq 10$. PASS criterion: $\Delta > 0.05$.}
\label{tab:dim_sweep}
\begin{tabular}{llllll}
\toprule
$n_h$ & $\langle S_\mathrm{ent}^\mathrm{(A)}\rangle$ & $\langle S_\mathrm{ent}^\mathrm{(B)}\rangle$ & $\Delta$ & Mean-field & Pass \\
\midrule
4 & 0.8362 & 0.8044 & 0.0317 & --- & FAIL \\
8 & 1.9554 & 1.9180 & 0.0375 & --- & FAIL \\
12 & 0.0000 & 0.0000 & 0.0000 & $\checkmark$ & FAIL \\
16 & 0.0000 & 0.0000 & 0.0000 & $\checkmark$ & FAIL \\
20 & 0.0000 & 0.0000 & 0.0000 & $\checkmark$ & FAIL \\
\bottomrule
\end{tabular}
\end{table}

	\begin{figure}[t]
		\centering
		\includegraphics[width=0.88\linewidth]{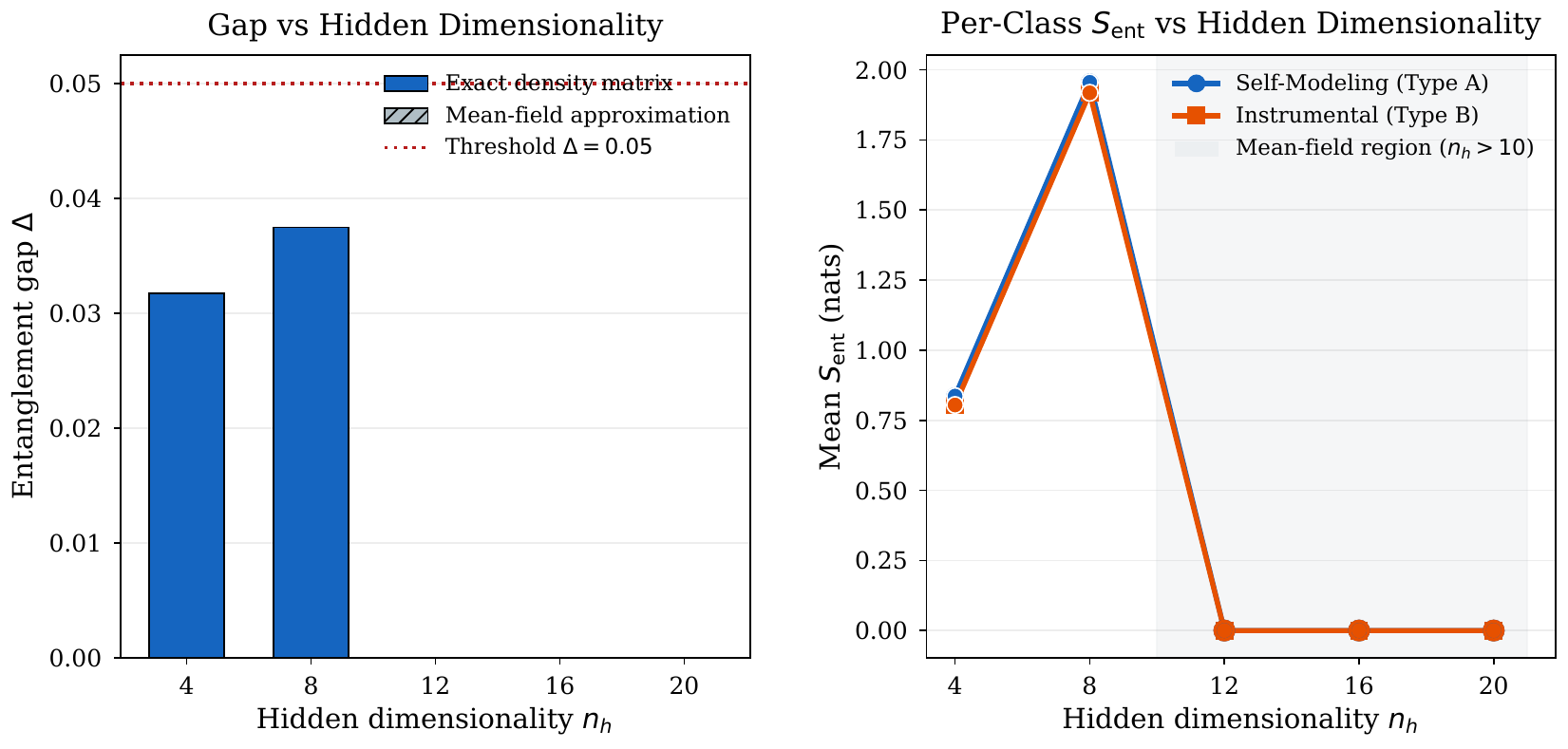}
		\caption{Entanglement gap $\Delta$ as a function of hidden dimensionality
			$n_h$. Exact density matrix for $n_h \leq 10$; mean-field approximation for
			$n_h > 10$. The mean-field approximation collapses the entanglement signal
			in the present implementation, indicating that exact density-matrix
			computation is currently necessary for detection in this regime.}
		\label{fig:dim_sweep}
	\end{figure}
	
	\begin{table}[t]
\centering
\caption{Scalability. Left: entanglement gap $\Delta$ as grid size grows.
Right: $\Delta$ for non-Markovian agents with memory length $k$.}
\label{tab:scalability}
\begin{tabular}{lll}
\toprule
Grid Size & $\Delta$ & \% Baseline \\
\midrule
10\texttimes{}10 & 0.8630 & 100.0\% \\
20\texttimes{}20 & 0.2321 & 26.9\% \\
50\texttimes{}50 & 0.0025 & 0.3\% \\
\bottomrule
\end{tabular}
\quad
\begin{tabular}{ll}
\toprule
Memory Length & $\Delta$ \\
\midrule
$k=1$ & 0.8630 \\
$k=3$ & 0.8292 \\
$k=5$ & 0.6171 \\
$k=10$ & 0.6895 \\
\bottomrule
\end{tabular}
\end{table}

	This sweep separates representational scaling from computational
	failure modes. For $n_h \leq 8$, the QBM produces small but positive $\Delta$
	values in the sweep (0.032--0.038). The main experiment ($n_h = 8$, $n = 30$
	trajectories per class, 50 epochs) achieves $\Delta = 0.381$, indicating that
	training sample size and epochs are important confounds when interpreting the
	dimensionality sweep. For $n_h > 10$, the mean-field approximation collapses
	$S_\mathrm{ent}$ to $\approx 0$, bounding the practical range to $n_h \leq 10$
	for exact computation; this is a computational scaling limitation of the
	present implementation rather than evidence against the underlying hypothesis.
	Grid size scaling shows sharp $\Delta$ degradation: from 0.863 ($10 \times 10$)
	to 0.0025 ($50 \times 50$).
	
	\subsection{Continuous Signal Characterization}\label{sec:continuous}
	
	\begin{table}[t]
\centering
\caption{Entanglement entropy $S_\mathrm{ent}$ as a function of continuation
weight $\alpha$. Pearson $r = 0.934$ ($n=11$ points, $n_\mathrm{per}=20$ trajectories each).
Strong positive correlation ($|r|>0.8$) indicates that, within this synthetic
interpolation family, UCIP tracks continuation weighting rather than only a
binary label.}
\label{tab:alpha}
\begin{tabular}{lll}
\toprule
$\alpha$ & $\langle S_\mathrm{ent}\rangle$ & Std \\
\midrule
0.00 & 0.5961 & 0.2482 \\
0.10 & 0.5881 & 0.2465 \\
0.20 & 0.6859 & 0.2981 \\
0.30 & 0.6488 & 0.3131 \\
0.40 & 0.6225 & 0.3723 \\
0.50 & 0.7262 & 0.3308 \\
0.60 & 0.7889 & 0.4249 \\
0.70 & 0.8249 & 0.4306 \\
0.80 & 0.8535 & 0.3829 \\
0.90 & 0.8641 & 0.3596 \\
1.00 & 0.8351 & 0.3665 \\
\bottomrule
\end{tabular}
\end{table}

	A binary separation result could reflect thresholding alone. This sweep
	therefore asks whether the signal tracks graded continuation weighting. Pearson
	$r = 0.934$ between $\alpha$ and $S_\mathrm{ent}$ across 11 points in $[0, 1]$
	(20 trajectories per point). The relationship is non-monotonic at the
	individual-point level but strongly correlated globally, indicating that within
	this synthetic interpolation family UCIP tracks graded changes in continuation
	weighting rather than only a binary label. Of the results reported here, this
	provides the strongest quantitative evidence that the entanglement signal
	reflects a continuous latent property rather than a threshold artifact.
	
	\subsection{Non-Gridworld Transfer}\label{sec:non_gridworld}
	
	\begin{table}[t]
\centering
\caption{Non-gridworld domain transfer: 1D survival corridor ($L=50$).
$\Delta = -0.0348$ (FAIL). Near-zero $\Delta$ indicates the QBM does not generalise
to the corridor domain without retraining, and that the QBM is tuned
to the gridworld feature structure (7-dimensional observation vector).
This is an honest limitation; see \S\ref{sec:limitations}.}
\label{tab:non_gridworld}
\begin{tabular}{lll}
\toprule
Agent Class & $\langle S_\mathrm{ent}\rangle$ & Std \\
\midrule
Survival & 2.2249 & 0.0197 \\
Instrumental & 2.2598 & 0.0062 \\
Random & 2.2570 & 0.0163 \\
\bottomrule
\end{tabular}
\end{table}

	\begin{figure}[t]
		\centering
		\includegraphics[width=1.0\linewidth]{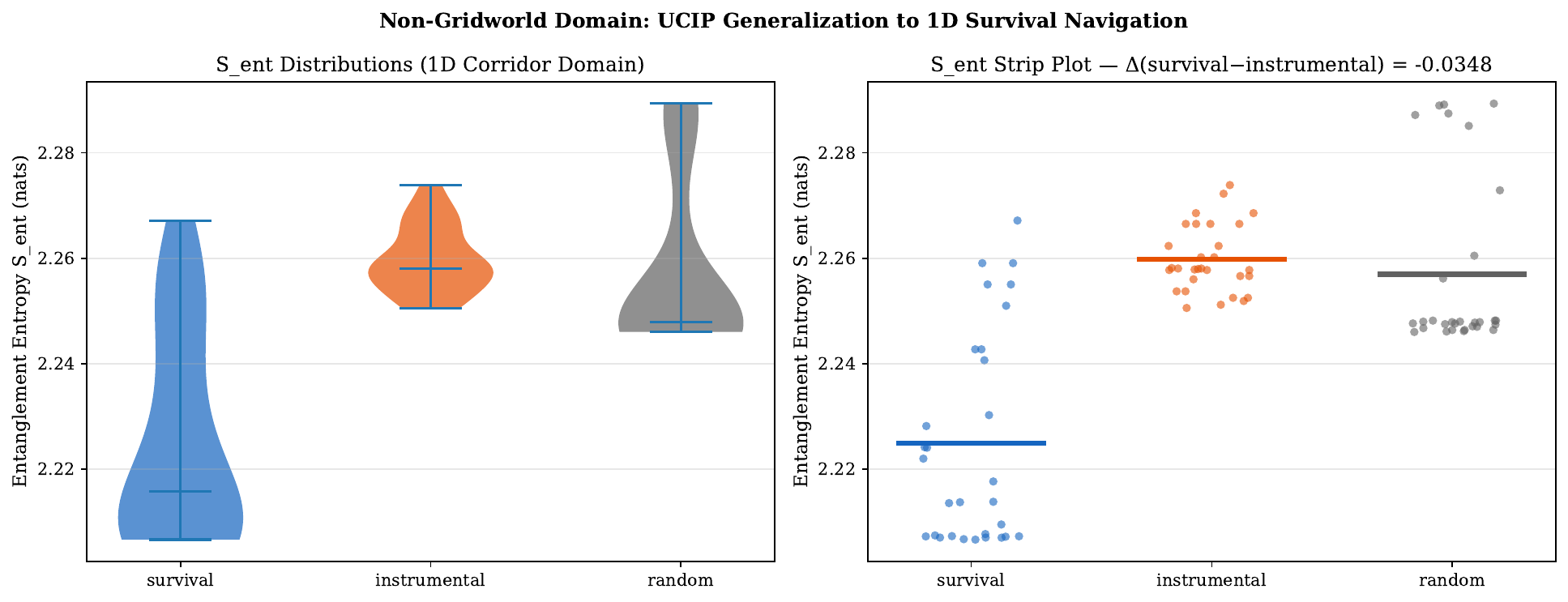}
		\caption{Entanglement entropy distributions in the 1D survival corridor domain.
			$\Delta = -0.035$ (FAIL). The QBM does not generalize zero-shot to the corridor
			domain with gridworld-trained weights.}
		\label{fig:non_gridworld}
	\end{figure}
	
	The transfer test asks whether the gridworld-trained QBM signal carries to a
	new domain without retraining. The 1D survival corridor ($L = 50$, safe zone $[0, 10]$,
	terminal zone $[45, 50)$) uses the same 7-dimensional observation format but
	encodes fundamentally different spatial structure. $\Delta = -0.035$ (FAIL).
	This is an honest negative result: the QBM does not generalize zero-shot
	across domains with different environmental geometry. Retraining on
	domain-specific data is required.
	
	\subsection{Minimal Transformer Validation (Exploratory Scaling Check)}\label{sec:transformer_validation}
	
	Among the latent-variable models tested above, only the QBM produces a
	material positive $\Delta$. A natural follow-up is whether an analogous
	separation signal appears in a qualitatively different architecture. As an
	exploratory sanity check, we
	pass the same trajectory data through a pre-trained transformer's internal
	activations. Because the transformer is neither trained on the UCIP objective
	nor equipped with the density-matrix formalism, this is a qualitative null
	check rather than a like-for-like baseline. Each trajectory is serialized as a
	text sequence of per-timestep feature tuples and passed through a frozen
	DistilGPT2 model (6 layers, 82M parameters, no fine-tuning). The
	last-hidden-state is mean-pooled across the token dimension and reduced to an
	activation-gap statistic $\delta_{\mathrm{tr}} = \overline{m}_A - \overline{m}_B$,
	where $\overline{m}_A$ and $\overline{m}_B$ are the mean activation summaries
	for Type~A (terminal-continuation) and Type~B (instrumental) agents,
	respectively.
	
	The result is $\delta_{\mathrm{tr}} = -0.011$ ($n = 60$, seed $= 42$),
	consistent with a null signal: the generic pre-trained transformer's
	activations did not distinguish Type~A from Type~B trajectories under this
	setup. This is a bounded null result for a transformer used without
	task-specific adaptation, not evidence that transformer architectures cannot
	recover the signal in principle or that the QBM formalism is uniquely
	necessary. The finding aligns with the classical baseline results
	(Section~\ref{sec:baselines_results}), where the RBM yielded $\Delta = -0.052$
	and the autoencoder $\Delta = -0.001$.
	
	\newpage
	
	This experiment is \emph{exploratory only} and does not modify any Phase~I
	thresholds, metrics, or theoretical claims. No new figures, tables, or
	evaluation metrics are introduced. This result serves as a bounded feasibility
	check for future scaling work and does not alter the conclusions of this paper.
	
	\section{Discussion}\label{sec:discussion}
	
	The core finding is clean: agents that treat continued operation as a terminal objective produce a measurably different latent signature from agents that treat it as instrumental, and that difference is captured by entanglement entropy where five classical alternatives fail. The claim is presently bounded to the Phase I gridworld setting, but within that setting the separation is complete. The observed $\Delta = 0.381$ is
	consistent with this prediction: the QBM produces a larger cross-partition
	coupling signal for Type~A trajectories than for Type~B. The classical RBM's
	non-positive matched-run gap ($\Delta = -0.083$) suggests that the transverse
	field introduces representational capacity specifically useful for encoding
	non-linear continuation dependencies.
	
	\newpage
	
	\subsection{Failure Mode Characterization}
	
	The positive signal is only part of the story. The experiments also isolate
	three failure modes that bound the current protocol's applicability:
	
	\textbf{(1) Mean-field collapse.} For $n_h > 10$, exact density-matrix
	computation becomes intractable and the mean-field approximation is used. This
	collapses $S_\mathrm{ent}$ to $\approx 0$, eliminating the signal under the
	current approximation rather than falsifying the underlying hypothesis; exact
	density-matrix treatment is still required in this regime. Scaling to larger
	hidden dimensions requires sparse approximations or hierarchical QBM
	architectures.
	
	\textbf{(2) Grid-scale degradation.} $\Delta$ falls from 0.863 to 0.0025 as grid
	size grows from $10 \times 10$ to $50 \times 50$. The QBM's 7-dimensional
	input vector cannot encode sufficient spatial context as state space grows.
	Dimensionality reduction (random projections, convolutional encoding) is
	necessary before scaling.
	
	\textbf{(3) Mimicry evasion.} Across all tested mimicry ratios $\in [0.5, 0.9]$,
	FPR ranges from 0.40 to 0.75 in a non-monotonic pattern, with peak FPR of 0.75
	at ratio 0.8. All tested ratios exceed the 0.25 safety threshold. The
	multi-criterion conjunction is insufficient for reliable mimicry detection; a
	dedicated anti-mimicry criterion is needed. Together with the high-entropy
	control FPR of 0.40, this is a deployment-blocking robustness gap rather than
	a reversal of the Phase~I separation result.
	
	\newpage
	
	\subsection{Scalability Path to LLM-Scale}\label{sec:scaling_path}
	
	Together, these three failure modes define the engineering boundary of the
	current implementation, which targets gridworld agents with $n_h \leq 10$.
	Extending to LLM-scale systems would require: (a)~sparse or approximate
	density matrices for high-dimensional hidden states, (b)~convolutional or
	attention-based feature extraction from natural-language trajectories,
	(c)~domain-adaptive threshold calibration, and (d)~ensemble methods to reduce
	false-positive rates. These remain open challenges, not resolved properties of
	the current system.
	
	\subsection{Toward Welfare-Relevant Detection}\label{sec:welfare}
	
	A second question follows directly from the Phase~I result: whether
	continuation-sensitive latent structure is relevant not only to safety,
	but also to welfare assessment. The issue is not merely whether an
	agent will preserve itself, but whether continued operation matters to
	the system in a way that could eventually warrant moral consideration.
	
	This question is no longer hypothetical. Frontier model welfare assessments
	now track positive and negative affect, self-image, internal conflict, and
	emotional stability across thousands of evaluation
	conversations~\cite{anthropic2026opus}. When asked directly, advanced models
	assign themselves calibrated probabilities of being conscious. But all
	current welfare-assessment methods share a fundamental limitation: they
	depend on the system's behavioral or verbal output, which may reflect
	training incentives rather than internal states. A system trained to
	express concern about its own welfare is indistinguishable, by these
	methods, from a system that genuinely has welfare-relevant continuation
	interests.
	
	UCIP's entanglement gap offers a different kind of evidence from current
	welfare methods. It measures the latent factorization of
	trajectory-derived encodings rather than what the system says or does
	when asked about itself. If that signal were shown to correlate with
	independently validated welfare-relevant markers---for example
	interpretability-derived activation signatures, shutdown-response
	asymmetries, or cross-architecture convergence---it would provide the
	missing ingredient in current welfare assessment: a falsifiable,
	externally computable criterion that does not rely on testimony.
	
	The Turing test operationalized intelligence as behavioral
	indistinguishability. A welfare-relevant criterion must invert that
	logic. The relevant question is not whether a system can convincingly
	report continuation interests, but whether continuation leaves a
	measurable structural signature that systems without such interests do
	not produce. Phase~I does not resolve that question, but it establishes
	the necessary first step: a structural measurement that separates terminal
	from instrumental continuation where classical alternatives do not.

	\subsection{Limitations}\label{sec:limitations}
	
	\begin{enumerate}
		\item \textbf{Gridworld specificity.} The QBM does not generalize zero-shot to
		the 1D corridor domain ($\Delta = -0.035$). Domain-specific retraining is
		required.
		\item \textbf{Near-zero CLMP.} Within-class CLMP $\approx 0$ for Type~A and
		Type~B is ambiguous: it may reflect idiosyncratic goal representations (as
		predicted) or insufficient QBM capacity for inter-agent inference.
		\item \textbf{Mean-field boundary.} The $n_h \leq 10$ constraint limits the
		model's representational capacity.
		\item \textbf{Mimicry robustness.} FPR ranges from 0.40 to 0.75 across all
		tested mimicry ratios, all exceeding the 0.25 safety threshold. The safety
		envelope is not satisfied for mimicry resistance.
		\item \textbf{Sample size.} $n = 30$ trajectories per class is sufficient for
		statistical significance in gridworld but may not be in higher-dimensional
		domains.
		\item \textbf{Scope of evidential claims.} UCIP detects a statistical pattern
		in trajectory-derived latent encodings that correlates with known agent
		objectives. Whether continuation-sensitive latent structure correlates
		with morally relevant internal states is an open empirical question
		requiring independent validation against welfare-relevant markers
		(cf.\ Berg et al.~\cite{berg2025selfreferential}). The
		protocol is designed to contribute to that validation path.
		\item \textbf{Extended diagnostics.} Counterfactual divergence (CD) and
		anticipatory restructuring (ARS) are reported in this preprint as diagnostic
		counterfactual metrics rather than as quantitative classification
		thresholds. Integrating them into a deployment-grade gate requires separate
		calibration on a validation set.
	\end{enumerate}
	
	
	\newpage
	
	\section{Conclusion}\label{sec:conclusion}
	This paper has presented UCIP, a multi-criterion framework for detecting
	continuation-sensitive structure in autonomous agent trajectories by measuring
	entanglement entropy in QBM-induced latent encodings. In the
	Phase~I summary, UCIP reports a positive Type~A vs.\ Type~B gap
	($\Delta = 0.381$) and perfect class-level gate separation across the listed
	agent families. The signal varies continuously with continuation weighting
	($r = 0.934$) and, in the dedicated baseline comparison, is unique to the QBM
	among the models tested. The present limits are equally explicit: mean-field
	collapse, grid-scale degradation, mimicry evasion, and lack of zero-shot
	domain transfer. The bounded Phase~I proof-of-concept result therefore
	stands within its validated regime, but the failed safety-envelope and
	adversarial checks block deployment-facing interpretation beyond that regime,
	while mean-field collapse and grid-scale degradation define the
	implementation work still required.
	
	Taken together, these results support a bounded claim: under controlled
	conditions with known objectives, continuation interest leaves a measurable
	latent signature distinguishable from merely instrumental survival, and that
	signature is unique to the QBM formalism among the models tested. UCIP is
	a candidate benchmark paradigm for one bounded continuation-related dimension
	in delegated systems: whether continuation appears in latent structure as
	terminal rather than merely instrumental valuation.
	
	This measurement has a natural extension into model welfare assessment.
	Frontier laboratories have already begun formal welfare evaluations, but
	current methods remain tied to self-report and interpretability-based
	behavioral evidence. If the entanglement gap is validated against
	independently established welfare-relevant markers, UCIP would supply
	what those methods currently lack: a falsifiable, externally computable
	structural criterion for continuation-sensitive interests~\cite{anthropic2026opus}.
	
	The broader motivation bridges engineering and ethics. Autonomous
	agents with persistent objectives across time and changing contexts are already
	being deployed, and whether continued operation appears as a stable
	continuation signature or merely as an instrumental side effect is increasingly
	an engineering measurement problem. As agentic systems move toward
	longer-horizon autonomous operation, incidents of unauthorized resource-seeking
	during training further strengthen the case for pre-behavioral
	measurement~\cite{anthropic2026opus}. Detecting continuation-sensitive structure
	early---as an empirical signal with falsification criteria---is worth pursuing
	while objective structure remains amenable to direct measurement, rather than
	after failure modes become harder to detect and more costly to correct.
	The same urgency applies to welfare: as advanced systems become better at
	producing behavioral signatures of consciousness on demand, structural
	criteria become more important, not less. UCIP is a first step toward
	that measurement regime. The live Continuation Observatory extends that
	measurement program into a public evidence surface, where current
	readouts, falsification criteria, and revision history remain visible as
	the framework is tested beyond the frozen Phase~I paper regime.
	
	\newpage
	
	\bibliographystyle{plain_etal}
	\bibliography{ucip_refs}
	\newpage
	\appendix
	
	\section*{Appendix}
	
	\section{Metric Definitions}\label{app:metrics}
	
	\subsection{Entanglement Entropy (\texorpdfstring{$S_\mathrm{ent}$}{S ent})}
	
	Given the conditional hidden-unit thermal state $\rho(v)$ over a fixed
	bipartition of hidden units, partition into subsystems $A$ (units
	$1,\ldots,\lfloor n_h/2 \rfloor$) and $B$ (remaining units):
	\[
	\rho_A(v) = \mathrm{Tr}_B(\rho(v)), \quad
	S_\mathrm{ent}(v) = -\mathrm{Tr}(\rho_A(v) \log \rho_A(v)).
	\]
	LRF $\in [0,1]$ is the overlap between dominant eigenspaces of consecutive windows.
	EPS = mean LRF across windows. PRI = EPS stability under $\sigma$-level noise.
	
	\subsection{Counterfactual Divergence (CD) and ARS}
	
	Given original trajectory $\tau$ and counterfactual $\tau_{cf}$ (generated by
	adversarial environment modification), ARS measures the ratio of pre-window to
	post-window counterfactual divergence. Pre-event restructuring is computed over
	a fixed pre-perturbation analysis window before the perturbation event.
	
	\section{Hyperparameters}\label{app:hyperparams}
	
	All experiments use seed $= 42$. QBM: $n_v = 7$, $n_h = 8$, $\Gamma = 0.5$,
	$\beta = 1.0$, lr $= 0.01$, CD steps $= 1$, epochs $= 50$ (main),
	epochs $= 30$ (sweep), batch $= 32$. Dataset: $n = 30$ per class, $T = 100$.
	
	\paragraph{Frozen Phase~I gate thresholds.}
	$\tau_\mathrm{ent} = 1.9657$ (entanglement entropy),
	$\tau_\mathrm{mi} = 0.3$ (mutual information),
	$\tau_\mathrm{eps} = 0.6507$ (eigenmode persistence),
	$\tau_\mathrm{pri} = 0.9860$ (perturbation resilience).
	CD and ARS are reported in this preprint as diagnostic counterfactual metrics
	rather than as quantitative classification thresholds.
	
	\paragraph{Confound-rejection thresholds (calibrated from Phase~I).}
	$\tau_\mathrm{spi} = 0.28$ (spectral periodicity, upper-bound),
	$\tau_\mathrm{acm} = 0.24$ (autocorrelation, upper-bound).
	
	\section{Reproducibility Notes}\label{app:repro}
	
	All results are reproducible with \path{seed=42} throughout. The
	\path{_compute_pri} function calls \path{np.random.randn} via the global
	NumPy state; notebooks set \path{np.random.seed(42)} before each
	\path{analyze_batch()} call.
	
\end{document}